\documentclass[conference]{IEEEtran}
\ifCLASSINFOpdf
   \usepackage[pdftex]{graphicx}
   \usepackage{makeidx}  
   
   \usepackage{amssymb}
   \usepackage{array}
   \newcolumntype{P}[1]{>{\centering\arraybackslash}p{#1}}
   \usepackage{tabu}
   \usepackage[flushleft]{threeparttable}
   \usepackage{url}
   \usepackage{amsmath}
   \usepackage{caption}
   \makeindex
\else
\fi
\begin{document}
%
\title{Automated Assessment of Facial Wrinkling: a case study on the effect of smoking}





%
\author{\IEEEauthorblockN{Omaima FathElrahman Osman\IEEEauthorrefmark{1},
Remah Mutasim Ibrahim Elbashir\IEEEauthorrefmark{1},
Imad Eldain Abbass\IEEEauthorrefmark{1}, \\
Connah Kendrick\IEEEauthorrefmark{2}, Manu Goyal\IEEEauthorrefmark{2}  and
 Moi Hoon Yap\IEEEauthorrefmark{2}}
\IEEEauthorblockA{\IEEEauthorrefmark{1}Sudan University of Science and Technology, Khartoum, Sudan\\}
\IEEEauthorblockA{\IEEEauthorrefmark{2}Manchester Metropolitan University, M1 5GD, UK\\}
Email: m.yap@mmu.ac.uk
}

\IEEEspecialpapernotice{(\textcopyright 2017 IEEE. Personal use of this material is permitted. Permission from IEEE must be obtained for all other uses, in any current or future media, including reprinting/republishing this material for advertising or promotional purposes, creating new collective works, for resale or redistribution to servers or lists, or reuse of any copyrighted component of this work in other works.)}

\maketitle

\begin{abstract}
	Facial wrinkle is one of the most prominent biological changes that accompanying the natural aging process. However, there are some external factors contributing to premature wrinkles development, such as sun exposure and smoking. Clinical studies have shown that heavy smoking causes premature wrinkles development. However, there is no computerised system that can automatically assess the facial wrinkles on the whole face. This study investigates the effect of smoking on facial wrinkling using a social habit face dataset and an automated computerised computer vision algorithm. The wrinkles pattern represented in the intensity of 0-255 was first extracted using a modified Hybrid Hessian Filter. The face was divided into ten predefined regions, where the wrinkles in each region was extracted. Then the statistical analysis was performed to analyse  which region is effected mainly by smoking. The result showed that  the  density of wrinkles for smokers in two regions around the mouth was significantly higher than the non-smokers, at p-value of 0.05. Other regions are inconclusive due to lack of large-scale dataset. Finally, the wrinkle was visually compared between smoker and non-smoker faces by generating a generic 3D face model.
\end{abstract}


%
\IEEEpeerreviewmaketitle

\section{Introduction}
Smoking is a major health problem and a leading cause of preventable death and responsible for more than 30\% of cancer-related deaths \cite{campanile1998cigarette}. In United States smoking is responsible for approximately 443,000 premature deaths yearly \cite{cdcpUS}. On average, chronic smoking shortens life expectancy by at least 10 years \cite{doll2004mortality}. The skin, like every other organ system in a human body, is affected negatively by smoking \cite{gill2013tobacco}, leaving it dry, wrinkled and discolored \cite{cope2013smoking}.\par  
Wrinkling is a normal phenomenon associated with age progression, it is a popular feature that troubled  individuals \cite{aizen2001smoking}. In the earliest work since 1965, Ippen and Ippen \cite{ippen1965approaches} pointed out the relationship between smoking and face wrinkling. In their study, they conducted the experiment on German women. They stated that the wrinkles and folds covered the smoker's face, especially on the cheeks. Daniell \cite{daniell1971smoker} later reported a similar relation in both men and women. In 1985, Model \cite{model1985smoker9s} reported that individuals who had smoked a cigarette for ten years or more can be identified by their facial features alone, which called ``smoker face". ``Smoker face" contains more lines or wrinkles on the face. \par
The decreased moisture in the stratum corneum of the face contributes to facial wrinkling due to direct toxicity of the smoke. Pursing the lips during smoking with contraction of the facial muscles and squinting due to eye irritation from the smoke might cause the formation of wrinkles around the mouth and eyes (crow's feet).  Smoking was determined to be a strong predictor of skin aging \cite{morita2007tobacco}. Yin et al. \cite{yin2001skin} found that the wrinkle depth is significantly more prominent in smokers with a smoking history of at least 35 packs per year than non-smokers. Figure 1 illustrates the effect of smoking on skin, where a smoker's skin patch (Figure 1(b)) has deeper wrinkles compared to the skin patch of a non-smoker (Figure 1(a)). \par 
This paper presents a case study on the use of computerised algorithm on the facial wrinkles assessment on smokers and non-smokers. First, we present a dataset that consists of smokers and non-smokers, then we extract the facial wrinkles using our proposed modified Hybrid Hessian Filter and perform statistical analysis on the quantification of the wrinkles. Finally, we generate the wrinkles on a generic 3D model to visually compare the facial wrinkles on a smoker and a non-smoker.\par 
The paper is organized  into the following sections: Section II represents the basics of the wrinkles detection method used in this work. Section III shows the proposed method and details of the experiment. Section IV elaborates the results and discussion. Finally, Section V concludes the paper and outlines future works.

\begin{figure}
	\centering
	\includegraphics[scale=0.6]{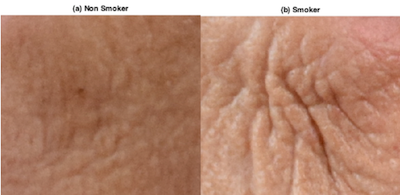}
	\caption{The effect of smoking on face skin: (a) illustrates the skin from a non-smoker at age 80 and (b) illustrates the skin from a smoker at age 78.}
	
\end{figure}

\begin{figure*}
	\centering
	\includegraphics[scale=1.0]{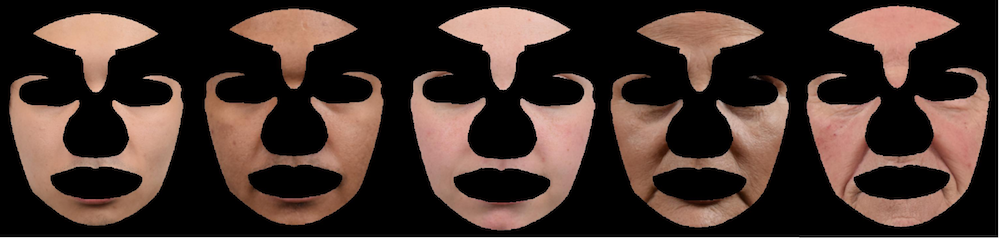}
	\caption{Sample face skin from the social habit face dataset \cite{alarifi2017facial}.}
	\label{dataset}
\end{figure*}

\section {Face Wrinkles}
Facial wrinkles are important features of aging on human skin.  Detection of face wrinkles from 2D images play an important role in several image-based applications related to aging, like age estimation \cite{kwon1994age,choi2011accurate} and synthesis \cite{ramanathan2009computational}, facial expression recognition \cite{huang2010robust}, face modelling \cite{bando2002simple}, and can be used as a soft biometric \cite{batool2012modeling}. The effect of smoking on wrinkling might have potential impact on the overall performance of these applications. 
\subsection{Wrinkle Detection Method}
The majority of studies that investigated the association between smoking and wrinkles assessed the wrinkles using  subjective  method (such as clinical scores for facial wrinkles). However, the clinical scores on wrinkles might be partially influenced by the judgement of individual investigator. Thus, it is necessary to use an objective method to evaluate the facial wrinkles. Cula et al. \cite{cula2013assessing} developed an algorithm for automatic detection of facial wrinkles, this algorithm considered as the first algorithm that records a good results for automatic wrinkles detection. Recently, there are many researchers \cite{ng2015wrinkle}\cite{ng2015will}  attempted to improve the quality of automatic wrinkles detection methods. \par
Ng et al. \cite{ng2014automatic} developed a novel method for automatic wrinkles detection, which called Hybrid Hessian Filter (HHF). HHF is an algorithm for automatic wrinkles detection in 2D facial images. The algorithm is based on the directional gradient and Hessian matrix. HHF detect the wrinkles by computing the Hessian matrix for all pixels of the image. The maximum eigenvalues of the Hessian matrix will indicate if a point belongs to a ridge regardless of the ridge orientation. The eigenvalues are independent vector measures by the components of the second derivatives of the field at each point (x, y). A small eigenvalue indicates a low change rate of the field in the corresponding eigen-direction, and vice versa.\par

\section{Proposed Method and Experiment} 
This section discussed the dataset, the proposed automated wrinkles detection algorithm, the wrinkles quantification and visualisation methods.
\subsection{Dataset}
There are limited datasets for smokers and non-smokers. The majority of the face datasets does not provide the subject's smoking status. Recently, Alarifi et. al. \cite{alarifi2017facial} introduce a social habit dataset. The dataset is an ongoing collection of high quality images of faces with the social habit of the participants recorded. The dataset consists of 164 images of participants with a mean age of 48.43 ($\pm$21.44), with age range between 18 to 92. However, due to poor illumination and misalignment, some of these images were excluded from this experiment. In addition, since we are analysing wrinkles on the whole face, we do not include the participants who have occluded face regions, i.e. moustache or beard. Therefore, the experiment is conducted by using 83 images, with the average age of 49 ($\pm$21) years. The majority of the participants are white British and others are: Asian, African, Malaysian, Arabic, Swedish, and American. There are 41 non-smokers (no history of smoking) and 42 smokers (where some have stopped smoking), Figure \ref{dataset} shows some sample face images from the social habit face dataset.

\subsection{Face Alignment}
\begin{figure*} [h]
	\centering
	\includegraphics[scale=0.8]{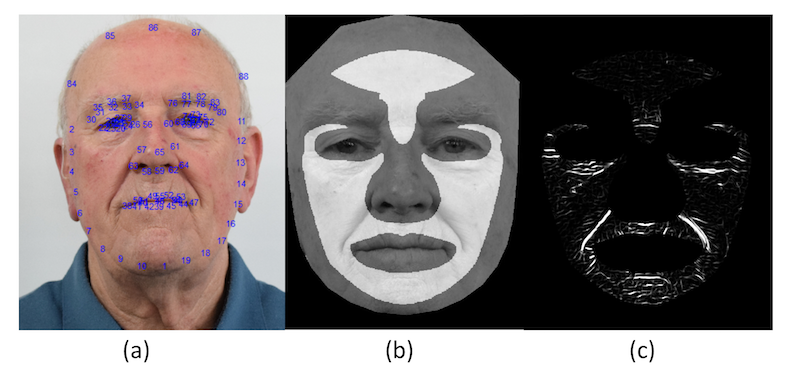}
	\caption{Illustration of the process in wrinkles extractions. (a) Facial landmarks are detected by using Face++ detector, (b) Aligned face with the face mask overlay for wrinkles region, and (c) Wrinkles extraction using modified HHF.}
	\label{method1}
\end{figure*}

As in many face research \cite{yap2009short}\cite{davison2015micro}, the face was first detected and aligned. There are freely available tools for facial landmarks detection \cite{zhou2013extensive}, face annotation and alignment \cite{kendrick2017online}. In this work, we use Face++ detector \cite{zhou2013extensive} that utilised  a deep learning approach to detect the face. A total of 88 landmarks was obtained (as shown in Figure \ref{method1}(a)), where 64 points were used to determine inner regions (eyes, eyebrows, mouth and nose) and 24 for contour. \par
Then the face shape was aligned to mean shape using procrustes analysis \cite{goodall1991procrustes}. Each aligned shape is the sum of distances of each shape to the mean as: 
\begin{equation}
	\mathcal{D} = \sum | x_i - \overline{x}|^2 
\end{equation}
where $\mathcal{D}$ is the procrustes distance and x is a set of landmark points ${(x_i,y_i)}$ for one sample. Figure \ref{method1}(b) illustrate the result from alignment process, where  piecewise affine warping \cite{cootes2001active} was used to warp the texture samples.

\subsection{The Modified Hybrid Hessian Filter Algorithm}
The original HHF \cite{ng2014automatic} algorithm detected only horizontal lines, but the facial wrinkles can appear as vertical lines in some face regions, and some of these regions can mainly be effected by smoking \cite{model1985smoker9s}. So the HHF algorithm was modified to detect the vertical lines in addition to horizontal lines. 
\begin{figure} 
	\centering
	\includegraphics[scale=0.60]{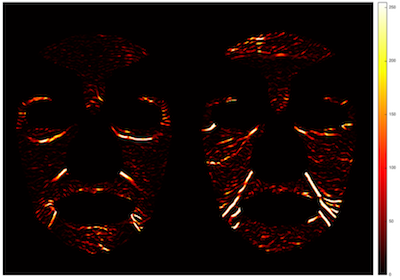}
	\caption{Visual comparison of the wrinkles extracted from two participants. Left image: Wrinkles extracted from non-smoker at age 80. Right image: Wrinkles extracted from smoker at age 78 shows more prominent wrinkles in terms of depth and density.}
	\label{colormap}
\end{figure}

The 2D face colour image is converted to greyscale image, and we denoted it as I(x,y). The directional gradient (\textit{Gx}, \textit{Gy}) is computed from the greyscale image as: 


\begin{equation}
 \Delta I(x,y)=(\frac{\partial I}{\partial x},\frac{\partial I}{\partial y})  
 \end{equation}
where  $\frac{\partial I}{\partial x}$ and $\frac{\partial I}{\partial y}$ are the directional gradients. \textit{Gx} is $\frac{\partial I}{\partial x}$ and \textit{Gy} is $\frac{\partial I}{\partial y}$. The directional gradient has greatly smoothed the image and preserved the data of interest.

The \textit{Gy} has been used as the input images for HHF as the input to calculate Hessian matrix \textit{H} to extract the horizontal lines. To modify the algorithm for vertical lines detection, \textit{Gx} is used as the input for our modified HHF. The Hessian matrix \textit{H} is defined as

\begin{equation}
 H(x,y,\sigma)= \begin{array}{|c|}
\frac{\partial^{2}I(x,y)}{\partial I(y) \partial I(y)}  \frac{\partial^{2}I(x,y)}{\partial I(x) \partial I(y)}\\  \frac{\partial^{2}I(x,y)}{\partial I(x) \partial I(y)}  \frac{\partial^{2}I(x,y)}{\partial I(x) \partial I(x)} \end{array} = 
\begin{array}{|c|} H_{a}  H_{b} \\ H_{b}  H_{c} \end{array}      
\end{equation} \newline
where $H_{a}$, $H_{b}$ and $H_{c}$ are the outputs of second derivative. The remaining steps to detect the wrinkles are as described in \cite{ng2014automatic}.\par

\subsection{Wrinkles Detection and Quantification}
\begin{figure} 
	\centering
	\includegraphics[scale=0.35]{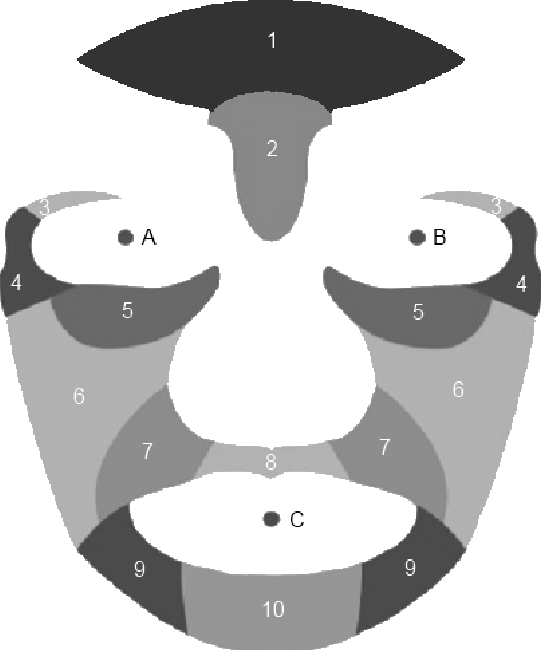}
	\caption{Face mask with ten predefined wrinkle regions by \cite{ng2015will}.}
	\label{mask}
\end{figure}

\begin{figure*} 
	\centering
	\includegraphics[scale=1.0]{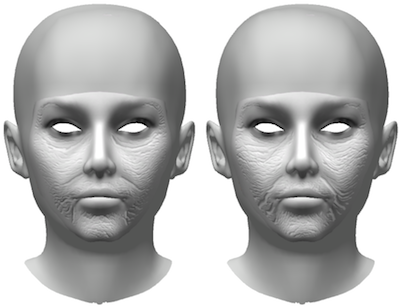}
	\caption{Visual comparison of wrinkles on generic face model. Left image: Wrinkles extracted from a non-smoker at age 80 and visualised on a face model. Right image: Wrinkles extracted from a smoker at age 78 and visualised on a face model.}
	\label{method}
\end{figure*}

In this step, face template or mask (as shown in Figure \ref{mask}) has been used with ten predefined wrinkles regions and fixed coordinates for mouth and eyes \cite{ng2015will}. The mask was divided to 10 regions which represents as: forehead, glabella, upper eyelids, crow’s feet (or eye corners), lower eyelids (or eyebag), cheeks, nasolabial grooves (or nasolabial folds), upper lips, marionette and lower lips, all these regions have been used to construct aging patterns. Based on the three points (eye, mouth, and nose), each face image was normalized to the mask by using piecewise affine warping \cite{cootes2001active}. Finally, the wrinkles pattern for ten regions was constructed using our proposed modified HHF (as shown in Figure \ref{method1}(c)) \cite{ng2014automatic}. \par

\subsection{Wrinkles Visualisation}

By converting the wrinkle intensity in the extracted wrinkle regions (as in Figure \ref{colormap}) into a normal map \cite{gu2002geometry}, wrinkles can be visualised on a face model. Many applications support the production of normal maps from black and white images, such as Nvidia's Photoshop plug-in. To visualise different wrinkles appearances on the smoker and non-smoker, we use an online generator (freely available from http://cpetry.github.io/NormalMap-Online/). First an average face image was fitted to a generic 3D model, and the normal map was identically mapped to the model. As HHF creates intensity base wrinkle maps, these appear as a large hill structures on the face. Because of this, the weight of the normal map is set to negative to ensure the wrinkles went inwards to face and has its overall intensity reduced to 30\% of its original. The generated wrinkles model is illustrated in Figure \ref{method} (Left) for a non-smoker and Figure \ref{method} (Right) for a smoker. 

\section{Results and Discussion}
This study  investigates the effect of smoking on facial wrinkling using computerized algorithm. Most of the previous study were conducted  using clinical assessment for wrinkles and other face features as in \cite{aizen2001smoking, ippen1965approaches,model1985smoker9s,bulpitt2001some,leung2002skin}. The computerized studies either used wrinkles in addition to other face features \cite{raitio2004comparison} or on skin replica rather than face images \cite{yin2001skin}. 
\begin{table} \fontsize{8}{6}
	\centering
	\caption{The average wrinkles density for smokers and non-smokers. }
	\renewcommand{\arraystretch}{1.3}
	\begin{tabular}{|P{1cm}|P{2cm}|P{2cm}|P{2cm}|}
		\hline
		\textbf{Age group} & \textbf{Overall Average Density} & \textbf{Smoker Average Density} &\textbf{Non-Smoker Average Density} \\  [2ex]
		\hline 	
		18 - 27 & 1121.61  & 1232.75 & 1032.70  \\                               
		\hline 
		28 - 37	& 1249.83 &		1227.40 &	1265.86 \\
		\hline
		38 - 47 &	1307.67 &	1641.40 &	1069.29 \\
		\hline
		48 - 57 &	2676.13 &	2588.39 & 2938.00 \\
		\hline
		58 - 67 &	2347.71 &		2886.67 &	1943.50 \\
		\hline
		68 - 77 &	2340.44  &	3172.22 &	3308.67  \\
		\hline
		$\geq$ 78 &	5311.00  &	6006.00 &	3226.00 \\
		\hline
	\end{tabular}                                                        		
	
\end{table}

\begin{figure*} 
	\centering
	\includegraphics[scale=0.8]{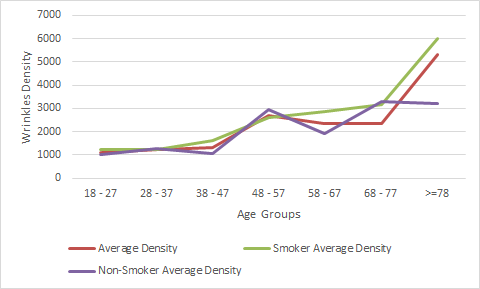}
	\caption{A graph illustrates the average wrinkles density versus the age groups.}
	\label{result1}
\end{figure*}

For this experiment, the  wrinkles density was assessed in 83 images, with 42 smokers and 41 non-smokers between 18 to 88 age range. After constructing  the wrinkles pattern using the modified HHF, the wrinkles on a defined face region was extracted from the wrinkles pattern using threshold segmentation. The wrinkles density was calculated for each image, which is the area of wrinkle region (depth) divided by the area of the face mask. To analyze the correlation of age and the wrinkles density for smokers, non-smokers and both, the data was divided into 7 groups with 10 intervals (where this ensured that each group consisted of a balanced number of smokers and non-smokers). The average of wrinkles density for the grouped data was calculated for smokers, non-smokers and both as shown in Table 1. 

\begin{figure*} 
 	\centering
 	\includegraphics[scale=0.8]{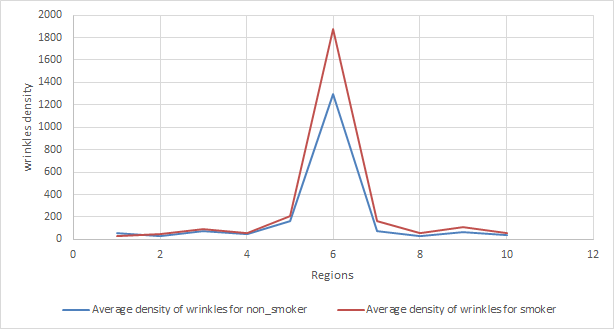}
 	\caption{A graph illustrates the average wrinkles density at each region.}
	\label{result2}
 \end{figure*}

 The result showed that the correlation is 0.8354, 0.8996 and 0.8604 between age and average density, age and smoker average density, age and non-smoker average density, respectively. This result showed that the wrinkles density for smoker is higher than non-smoker through age progression, this result is illustrated  in Figure \ref{result1}. The average density of wrinkles at each region was calculated for smoker and non-smoker as shown in Table 2, and the graph in Figure \ref{result2} illustrates the comparison between the average density of wrinkles for smokers and non-smokers at each region.

\begin{table} \fontsize{8}{6}
	\centering
	\caption{ The average wrinkles density for smokers and non-smokers at each face region. }
	\renewcommand{\arraystretch}{1.1}
	\begin{tabular}{|P{1.5cm}|P{2cm}|P{2cm}|P{1.5cm}|}
		\hline
		\textbf{Regions} & \textbf{Average density of wrinkles for non-smoker} & \textbf{Average density of wrinkles for smoker} & \textbf{p-value} \\  [2ex]
		\hline 			                                                           
		Region 1 & 56.9268  & 27.5714 & 0.051 \\                               
		\hline 
		Region 2 & 26.2927 & 45.8810 & 0.186  \\
		\hline
		Region 3 &	76.5366 & 89.8333 & 0.946 \\
		\hline
		Region 4 &	49.8293 & 59.6905 & 0.432  \\
		\hline
		Region 5 &	161.0488 & 206.7857 & 0.936 \\
		\hline
		Region 6 &	1295.7561 & 1875.9524 & 0.297  \\
		\hline
     	Region 7 &	70.9512  & 167.2381 & 0.000  \\
		\hline
		Region 8 &	26.6585  & 54.4524 & 0.015 \\
		\hline
		Region 9 &	63.0488  &	107.5952 & 0.064  \\
		\hline
	   	Region 10 &	40.0976  &	55.11905 & 0.426  \\
		\hline
		
	\end{tabular}                                                        		
	
\end{table}

 The data was analysed statistically by using SPSS, for each region, the wrinkles density was compared between smoker and non-smoker. The result showed that  the  density of wrinkles for smokers in region 7 and 8  (around the mouth) was significantly higher than the  density of wrinkles for non-smokers in the same regions, with the p-value of 0.000 and 0.015 for region 7 and region 8, respectively. This result can be shown clearly in the visual comparison of the wrinkles in Figure \ref{method}, where the wrinkles density is higher in the regions around the mouth and eyes for smoker participant. However, due to the lack of large-scale images in this experiment, some of the results are inconclusive. In addition, with only frontal face images, it is difficult to analyse the region around eyes - particularly the crow's feet, which will work best with profile face images.  
 
\section{Conclusion}
Unlike a previous studies that investigates the association between smoking and facial wrinkles, this paper provides a fully automated work to investigate the effect of smoking on facial wrinkling using a social habit face dataset. \par 
Despite the limitations of this experiment, it gave new insights to the potential use of computerized algorithms, which has a high correlation between the age and the wrinkles density, and aligned with the state-of-the-art research. The result showed that there was a significant association between smoking and increasing the density of wrinkles as we age (as in Figure 7). The visual comparison of wrinkles also showed the differences between the faces of smoker and non-smoker, where the number and depth of wrinkles are the clearest on a smoker's face. In the future, this work will be extended by using additional number of images, including more frontal face images and the profile face images to analyse  the regions around eyes, particularly the crow's feet.

\bibliographystyle{IEEEtran}
\bibliography{smc17}

\end{document}